\documentclass[authoryear,preprint,review,12pt]{elsarticle}

\usepackage{graphicx}
\usepackage{amssymb}

\usepackage{caption}
\usepackage{subcaption}

\usepackage{lineno}

\usepackage[utf8]{inputenc}     
\usepackage[T1]{fontenc}        
\usepackage{booktabs}			
\usepackage{threeparttable}		
\usepackage{multirow}
\usepackage{siunitx}
\usepackage[export]{adjustbox}
\usepackage{tabularx}
\usepackage[table]{xcolor}
\usepackage{color}              
\usepackage{hyperref}

\usepackage{adjustbox}

\journal{Engineering Applications of Artificial Intelligence}

\begin{document}

\long\def\/*#1*/{}


\title{A multi-head deep fusion model for recognition of cattle foraging events using sound and movement signals}



\author[add1]{Mariano~Ferrero}
\author[add1,add2]{José~O.~Chelotti}
\author[add1,add3]{Luciano~S.~Martinez-Rau}
\author[add1]{Leandro~D.~Vignolo}
\author[add4]{Martín~Pires}
\author[add4,add5]{Julio~R.~Galli}
\author[add1]{Leonardo~L.~Giovanini}
\author[add1,add6]{H.~Leonardo~Rufiner}

\address[add1]{Instituto de Investigación en Señales, Sistemas e Inteligencia Computacional, sinc(i), FICH-UNL/CONICET, 3000 Santa Fe, Argentina}
\address[add2]{TERRA Teaching and Research Center, University of Liège, Gembloux Agro-Bio Tech (ULiège-GxABT), 5030 Gembloux, Belgium}
\address[add3]{Department of Computer and Electrical Engineering, Mid Sweden University, Sundsvall, Sweden}
\address[add4]{Facultad de Ciencias Agrarias, Univ. Nacional de Rosario, S2125 Zavalla, Argentina}
\address[add5]{Instituto de Investigaciones en Ciencias Agrarias de Rosario, IICAR, Facultad de Ciencias Agrarias, UNR-CONICET, S2125 Zavalla, Argentina}
\address[add6]{Laboratorio de Cibernética, Facultad de Ingeniería, Univ. Nacional de Entre Ríos, Oro Verde 3100, Argentina}

\begin{frontmatter}
\begin{abstract}
Monitoring feeding behaviour is a relevant task for efficient herd management and the effective use of available resources in grazing cattle. The ability to automatically recognise animals’ feeding activities through the identification of specific jaw movements allows for the improvement of diet formulation, as well as early detection of metabolic problems and symptoms of animal discomfort, among other benefits. The use of sensors to obtain signals for such monitoring has become popular in the last two decades. The most frequently employed sensors include accelerometers, microphones, and cameras, each with its own set of advantages and drawbacks. An unexplored aspect is the simultaneous use of multiple sensors with the aim of combining signals in order to enhance the precision of the estimations. In this direction, this work introduces a deep neural network based on the fusion of acoustic and inertial signals, composed of convolutional, recurrent\textcolor{black}{,} and dense layers. The main advantage of this model is the combination of signals through the automatic extraction of features independently from each of them. The model has emerged from an exploration and comparison of different neural network architectures proposed in this work, which carry out information fusion at different levels. Feature-level fusion has outperformed data and decision-level fusion by at least a 0.14 based on the F1-score metric. Moreover, a comparison with state-of-the-art machine learning methods is presented, including traditional and deep learning approaches. The proposed model yielded an F1-score value of 0.802, representing a 14\% increase compared to previous methods. \textcolor{black}{Finally, results from an ablation study and post-training quantization evaluation are also reported.}
\end{abstract}


\end{frontmatter}


\section{Introduction}

\label{S:1}

The intensification of livestock production systems requires innovative tools to improve efficiency while mitigating environmental negative impacts. Traditional methods of livestock management, often based on herd-level observations, may overlook individual behavioral patterns, leading to suboptimal resource use and increased environmental footprints.

Individualized livestock monitoring offers significant economic benefits, including improved feed efficiency, reduced effluents, and enhanced animal health management \citep{Laca2009}. For instance, by accurately detecting foraging events, farmers can fine-tune feed distribution, ensuring that animals receive adequate nutrition without overfeeding. This precision not only \textcolor{black}{reduces feed costs \textendash ~a major} expense in livestock systems \textendash ~but also minimizes competition for limited resources. Furthermore, early detection of irregular behaviors through individual monitoring can aid in identifying health issues \citep{MorganDavies2024}, reducing veterinary costs and potential production losses.

From an environmental perspective, individualized monitoring contributes to sustainability by promoting optimal grazing practices. Overgrazing, a common issue in unmanaged systems, can lead to soil degradation, loss of biodiversity, and decreased carbon sequestration. By accurately identifying and managing foraging behavior, producers can implement rotational grazing strategies that enhance pasture resilience and soil health. Additionally, better feed management reduces greenhouse gas emissions per unit of production, aligning with global goals to mitigate climate change.

With regard to the traditional monitoring of feeding behaviour, two principal activities are considered: grazing and rumination. Despite this, a more fine-grained classification might be possible including drinking, chewing, foraging, \textcolor{black}{and} walking, among others \citep{Kilgour2012, Dasilvasantos2023}. Including these activities might contribute to provide a comprehensive analysis of feeding patterns, nutritional intake, and overall well-being. By accounting for these behaviors, a more complete understanding of the animal's feeding dynamics can be achieved.

Each period of the key activities mentioned before (grazing and rumination) may last from minutes to hours and consists of sequences of specific jaw movement (JM) events that allow their accurate identification and tracking.

These events are classified as bite, chew, and chew-bite (a combination of the two previous events) \citep{Laca2000-ac, Ungar2006-do, Milone2012-ap}. Monitoring the occurrence of these events and activity periods allows for the estimation of dry matter intake \citep{Galli2006-mv}, the detection of the presence of a disease or condition \citep{Calamari2014-ia, Paudyal2018-mv}, the prediction of states of stress \citep{Herskin2004-xv} or anxiety \citep{Bristow2007-bw}, and approximating the calving moment \citep{Buchel2014-pn, Clark2015-ts}, to name a few examples.

Continuous direct observation of cattle behaviours represents a challenge, especially when dealing with a significant number of animals distributed across extensive areas. This challenge has driven research into the use of sensors for monitoring relevant livestock behaviours. Various types of sensors have been proposed, allowing for differentiation between those which are positioned on the animal (commonly referred to as "wearables") and those situated externally. The former has been the predominant choice in the literature, with motion sensors being the preferred option, followed by acoustic sensors \citep{Chelotti2023-gk, Andriamandroso2016-ay}. 

Acoustic sensors are able to capture signals with high discriminative power, although the disadvantage is the difficulty in processing \textcolor{black}{them} due to the volume of generated information. On the other hand, the processing of IMU signals is simpler due to the smaller number of samples per second. Although these signals record important information about position, turns and other head movements, the discrimination of different JM events might be challenging \citep{Dasilvasantos2023, Chelotti2023-gk}.

While the use of a single sensor has been the most extensively studied approach, the combination of signals from multiple sensors has yet to be fully explored. This represents an advantage in this problem due to the ability to have complementary information to reduce environmental noise, make the system more robust to failures, and improve detection capabilities, among others. This promising approach can be addressed through the use of data fusion strategies combining the most used signals in the state-of-the-art: motion and audio signals.

In the context of information fusion, three main levels of abstraction are frequently employed in situations where data comes from multiple sensors. These are data fusion, feature fusion, and decision fusion \citep{Hall1997-eg, Qiu2022-qq}. Data fusion level refers to the premature combination of acquired signals from sensors to create a unique signal with several channels, regardless of whether pre-processing is performed or not. In this context, a common approach consists of the creation of multimodal signals by stacking raw signals. On the other hand, the feature-fusion level involves extracting representative values of each signal (usually using fixed-size windows) and then constructing a vector of fixed-dimension elements. The main idea is to combine information from all available signals in this single representation, generating some independence between specific properties of each signal \citep{Spinsante2016-sy}. Feature generation can be manual (i.e. following a feature engineering approach) or automatic (i.e. self-learned features in a deep learning approach). Finally, the decision-level fusion builds a system that combines predictions from underlying systems, each of which analyses information from a single sensor \citep{Garcia-Ceja2018-fq}. Consequently, the system endeavours to optimise the output by combining or selecting hypotheses generated by simpler systems, in accordance with a comparable methodology to ensemble methods \citep{Dietterich2000-ia}. To create a final decision, traditional approaches could be employed (such as majority voting) in addition to machine learning models (for instance decision trees or logistic regression).

This paper presents a multi-head convolutional neural network (CNN) - recurrent neural network (RNN) approach for the recognition of JM events in grazing cattle. The approach fuses information from acoustic and inertial measurement units (IMU) signals at the feature-level without any prior preprocessing or feature extraction. The proposed model is capable of detecting and classifying JM events simultaneously, distinguishing between five different classes. An investigation into the efficacy of different information fusion architectures has been conducted to identify the optimal configuration for enhancing recognition results in this context. Furthermore, the proposed method has been subjected to empirical evaluation and benchmarked against a range of state-of-the-art alternatives. Experiments were performed to show the superiority of multimodal approaches over unimodal solutions and to illustrate the advantages of deep architectures over traditional machine learning approaches. 

\textcolor{black}{An in-depth exploration of the technical details and implications involved in implementing the proposed model is beyond the scope of this study.}

The main contributions of this publication are the following:

a) It presents a multi-head CNN-RNN model that performs information fusion at the feature-level.

b) It proposes and evaluates different architectures of deep neural networks that perform data fusion at different levels.

c) It examines the effectiveness and accuracy of the proposed solution by comparing the obtained results with those obtained by state-of-the-art methods.

d) It presents an ablation study to analyse the benefits of each part of the proposed model.

Our proposed multi-head deep fusion model, leveraging sound and movement signals, provides a novel approach to detecting cattle foraging events with high precision. By integrating these modalities, our work addresses the gap in individualized livestock monitoring technologies and supports sustainable and economically viable livestock production systems.

The structure of the remaining parts of the article is as follows: Section 2 introduces a short overview of the state-of-the-art regarding automatic monitoring of ruminant feeding behaviour. Section 3 describes the proposed feature-level fusion model as well as other fusion level architectures proposed and analysed. Section 4 is dedicated to the experimentation including a description of the adopted methodology. Several comparisons are also presented in this section. Finally, conclusions, limitations, and future research lines are discussed in Section 5.

\section{Related work}
\label{S:2}

In the last \textcolor{black}{few} decades, ruminant feeding monitoring has attracted scientific attention due to the existing challenges and potential benefits from a practical point of view. Machine learning algorithms are proposed as a means of creating systems capable of working in this context. This section describes the recent developments in ruminant feeding monitoring analysing the most common sensing principles adopted.

\subsection{Sensors}
\label{S:2.1}

Motion sensors allow for the identification of specific ruminant behaviours based on changes in body posture. The principle of motion sensing and its location on the animal determines which movements can be monitored. Accelerometers have been the most studied sensor \citep{Aquilani2022-bj}, due to their low cost, compact size, and low power consumption \citep{Chelotti2023-gk}. Another advantage of the signals captured by this sensor is the low computational cost required for processing them, as they operate at sampling frequencies below 100 Hz. In the context of ruminant feeding monitoring, the use of motion sensors has been primarily focused on detecting activities such as rumination, grazing, and drinking \citep{Aquilani2022-bj}. However, their use for specifically detecting JM events poses challenges due to the limited discriminatory power of the signals captured for this purpose \citep{Chelotti2023-gk}. A variety of approaches have been explored, including the use of accelerometers \citep{Tani2013-ky, Oudshoorn2013-ez, Bloch2023-xm}, accelerometers and gyroscopes (referred to as IMUs) \citep{Andriamandroso2015-qf, Li2022-kp}, and accelerometers, gyroscopes, and magnetometers (referred to as inertial and magnetic measurement units) \citep{Liu2023-to}.

In free-grazing conditions, acoustic sensors have been demonstrated to be a valuable tool for monitoring feeding behaviour \citep{Ungar2006-do}. Microphones positioned on the animal's forehead are able to capture sounds produced by the teeth, transmitted through the bones, cavities, and soft tissues of the head \citep{Laca1992-gm, Galli2020-bm}. The information captured in these signals allows for the precise recognition of JM events \citep{Navon2013, Chelotti2018-ls, Martinez-Rau_2024}, as well as grazing and rumination activities \citep{Chelotti2023-ua}. However, the challenge in exploiting these signals lies in the presence of environmental noise and the computational requirements to process them \citep{Deniz2017}. Furthermore, the volume of information generated in a given time period is greater than that produced by motion sensors.

\subsection{Machine learning approaches}
\label{S:2.2}

With regard to the development of an automated system capable of classifying JM events and feeding activities, machine learning techniques have been extensively studied \citep{Chelotti2023-gk}. The most commonly used approaches follow a classic pattern recognition pipeline: pre-processing, feature extraction, and classification \citep{Bishop2006}. Nevertheless, certain limitations have been observed in the classification of JM events \citep{Chelotti2018-ls, Martiskainen2009-yl, Greenwood2017-gf} and feeding activities \citep{Chelotti2020-pj, Giovanetti2017-ow}. One of the principal limitations of these approaches is the necessity to manually specify the input features of the machine learning models. This aspect introduces a challenge in this problem because there is no consensus on which features should be employed \citep{Chelotti2023-gk}.

As an attempt to address this issue, within the field of deep learning, the use of CNNs has emerged. These architectures are capable of automatically learning features by adapting the filters or weights contained in the network. \citet{Li2021-wu} evaluated the use of CNNs on time-frequency representations of acoustic signals to classify JM events in dairy cows. The reported results are comparable or superior to those obtained through traditional schemes \citep{Chelotti2016-pf}. \citet{Wang2021-jm} explored the use of different deep neural network architectures to classify JM events in sheep from audio files. The proposed approach detects JM events using a heuristic method and subsequently performs classification using deep neural networks. Specifically, the use of fully-connected neural networks (FNNs), CNN, and RNN is evaluated. The input to the CNN and RNN is obtained by calculating Mel-frequency cepstral coefficients. In the case of the FNN, the input data consists of the raw signal corresponding to the previously detected event. \citet{Ferrero2023-vr} proposed a full end-to-end approach which combines FNN, CNN, and RNN to recognise JM events from acoustic signals. The model input constitutes signal chunks extracted using fixed-length time windows. The comparison with other state-of-the-art methods demonstrated a clear improvement over traditional approaches. \citet{Nunes2021} presented a similar approach using RNN to classify JM events in horses from acoustic signals with promising results. The use of deep neural networks has also been applied to inertial signals in the context of recognising feeding activities \citep{Peng2019-hv, Pavlovic2021-jx, Wu2022-mp, Bloch2023-xm}, with promising results.

Architectures that have yielded very good results in related problems such as attention mechanisms \citep{Topaloglu2023, Aydogmus2023}, have not been applied in this context. One explanation for this may be due to the scarcity of labeled data, which may be an impediment to train models with these characteristics.

\subsection{Multimodal learning outside JM events recognition}
\label{S:2.3}

The utilisation of independent sensing principles for the monitoring of feeding behaviour has been extensively addressed. However, the integration of diverse complementary information sources to achieve more robust and scalable performance in dynamic real-world environments is a promising and underexplored area of study \citep{Chelotti2023-gk}. The use of multimodal systems has been demonstrated to be beneficial in other areas, including speech recognition \citep{Mroueh2015-mo}, emotional state recognition \citep{Tzirakis2017-mp}, and human activity recognition \citep{Nweke2019-dr}.

\citet{Arablouei2023-lg} proposed a method that combines an accelerometer with global navigation satellite system (GNSS) data to classify feeding activities in cows. The solution involves first extracting a set of features from inertial signals and another set from GNSS signals. Subsequently, information fusion is explored at \textcolor{black}{the} feature and decision level. A FNN was used to construct the classification model. The reported results demonstrate that information fusion leads to superior outcomes compared to unimodal systems.

The evidence presented in this section indicates the existence of an untapped potential for enhancing JM events recognition. This potential \textcolor{black}{is} based on the utilisation of multimodal signals, which allows the exploitation of the advantages offered by each sensing principle. Furthermore, another aspect that has not been studied thus far is the generation of deep learning architectures capable of merging these signals and autonomously learning features, subsequently enabling the recognition of the JM events present in them. Results reported in the literature \citet{Ferrero2023-vr} suggest that the combination of convolutional and recurrent architectures emerges as a promising line of research on this problem.

\section{Methodology}
\label{S:3}

This section describes a multimodal deep learning architecture based on the combination of three types of neural networks: CNN \citep{Lecun1998-cb}, RNN \citep{Rumelhart1986-kj} and FNN \citep{Bishop2006}. In the following, a brief introduction to these architectures is provided. Then, a detailed description of the proposed method is presented with other proposed architectures which perform fusion at different levels are also introduced. Lastly, the dataset used in the experimentation is described.

\subsection{CNN, RNN and FNN}
\label{S:3.1}

FNN refers to a traditional neural network architecture in which each node belonging to a layer is connected with all nodes of the previous layer. This architecture has been used in classification and regression problems \citep{Bishop2006}. There are usually three types of layers including input, hidden, and output layers. While the neurons of the input layer represent the features provided to the network (input data or outputs from other networks), each neuron of the hidden and output layers represents a processing element that combines the output of incoming connected neurons using a non-linear activation function. The overall formal representation for a single hidden layer network is expressed in Eq. \ref{eq:eq1}.

\begin{equation}\label{eq:eq1}
y_{k}(x, w) = \sigma \left ( \sum_{j=1}^{M} w_{j_{i}}^{(2)} h \left ( \sum_{i=1}^{D} w_{j_{i}}^{(1)} x_{i} + w_{j_{0}}^{(1)} \right ) + w_{k_{0}}^{(2)} \right )
\end{equation}

Herein, $y_k$ denotes the output of the neuron $k$ based on the input vector $x$ of size $D$ and a set of weights $w$, $h$ denotes the activation function of $M$ neurons in the hidden layer, whereas $\sigma$ represents the activation function of the output neuron. The strength of the connections between neurons in FNNs ($w$ in Eq. \ref{eq:eq1}) is controlled using weights, which are optimised during the training process to adapt the model outputs to a set of desired values \citep{Bishop2006}.

CNNs \citep{Lecun1998-cb} are one of the most widely used architectures in recent decades. These networks usually consist of several convolutional layers, and each layer contains one or more filters (a set of arbitrary decimal numbers) to produce an output feature map of its inputs. In the learning stage, the weights of the filters (used in traditional convolutional mathematical operations) are adjusted to approximate the outputs using optimisation strategies as described above for FNNs. By doing this, the layers are capable of learning different high- and low-level patterns without explicit domain knowledge. In the field of information fusion, several sub-models (usually referred to as heads) could be independently applied to input signals to extract relevant features from them. In the case of a one-dimensional (1D) CNN with $n$ heads, the expression of the output value $z$ at position $i$ in feature map $m$ at layer $l$ of head $c$ can be denoted by Eq \ref{eq:eq2}.

\begin{equation}\label{eq:eq2}
z_{i}^{c l m} = h \left ( \sum_{j=0}^{F-1} x \times w_{j}^{clm} \right )
\end{equation}

Here, $h$ indicates the activation function for the kernel of size $F$ and weights $w_j$, and $x$ represents the signal affected by the kernel. 

In CNNs, convolutional layers are complemented by other types of layers, such as pooling, batch normalisation, and dense layers. Pooling layers perform simple mathematical operations on patches of the feature maps, such as extracting the maximum value, to reduce the dimensionality of the input. Batch normalisation layers, on the other hand,  perform input standardisation to speed up the network training process. Dense layers are equivalent to hidden layers in FNNs and allow the network to adapt the intermediate representations learned by the convolutions to effectively influence the final output. The connection between convolutional and dense layers is established by a flattening operation to convert the output of the convolutional layers into a 1D vector. 

Although FNNs can be used in problems with sequential or time series data, they present certain challenges that make them inappropriate in these scenarios. To address this limitation, RNNs emerged \citep{Rumelhart1986-kj}. In this architecture, layer outputs are connected as inputs to the same layer. A variation of \textcolor{black}{an} RNN known as Bidirectional RNN \citep{Schuster1997-uj} \textcolor{black}{adds a copy} of the proposed network trained on the reverse data sequence. Both independently trained RNNs are then connected to the next layer of the network.

Early RNN architectures have certain drawbacks related to the ability to learn efficiently from long sequences and new alternatives have been proposed. Gated recurrent units (GRUs) are a type of RNN in which each neuron has two different gates: reset and update \citep{Cho2014-zo}. These gates control how much information from previous and current states is used. A GRU architecture, in contrast with simple RNNs, effectively \textcolor{black}{captures} long-term dependencies in sequences by addressing the vanishing gradient problem. Additionally, GRUs are computationally more efficient and require fewer parameters than \textcolor{black}{Long Short-Term Memory \citep{Hochreiter1997}, another RNN type which includes three gates}, making them faster to train while still providing improved performance over simple RNNs, especially in tasks requiring memory of long-term dependencies.

\begin{figure}
       \centering
       \includegraphics[scale=0.8]{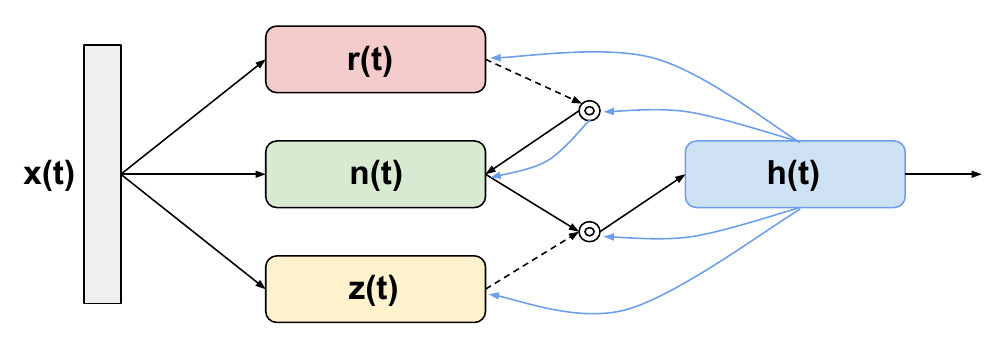}
       \caption{GRU cell diagram including the different gates and their connections.}
       \label{fig:GRU}
\end{figure}

A representation of a GRU cell is shown in Figure~\ref{fig:GRU},  and the associated mathematical expression is given in Eq.~\ref{eq:eq3} to \ref{eq:eq6}.

\begin{equation}\label{eq:eq3}
r_{t} = \sigma \left ( W_{r} x_t + W_r h_{t-1} + b_r \right )
\end{equation}

\begin{equation}\label{eq:eq4}
z_{t} = \sigma \left ( W_{z} x_t + W_z h_{t-1} + b_z \right )
\end{equation}

\begin{equation}\label{eq:eq5}
n_{t} = \phi \left ( W_n + r_t \odot \left ( W_n h_{(t-1)} \right ) + b_n \right )
\end{equation}

\begin{equation}\label{eq:eq6}
h_{t} = (1 - h_z) \odot n_t + z_t \odot h_{(t-1)}
\end{equation}

Herein, $x_t$ represents the input vector, $h_t$ the output vector, and $z_t$, $n_t$ and $r_t$ are the update, new and reset gate vectors, respectively at time $t$. $\sigma$ and $\phi$ represent the activation functions, whereas $W$ and $b$ are the parameters matrices and the bias vector of each gate, respectively. Bidirectional GRUs (BGRUs) have shown promising results in sound events detection \citep{Yihan2021-tk} and classification \citep{Zhu2020-cb}.

Stochastic gradient descent and backpropagation \citep{Rumelhart1986-kj} are very common algorithms to perform parameter optimisation in neural networks. In this context, artificial neural networks tend to overfit training data. To reduce the possibility of this, a dropout operation is used. This regularisation technique introduces random cuts between layer connections during training \citep{Hinton2012-bb}.

\subsection{Proposed model architecture}
\label{S:3.2}

Several deep neural network architectures could be proposed to merge the available acoustic and motion signals in this problem. Here, an architecture has been chosen that is capable of extracting features from each signal independently and combining them into a common feature space (feature-level fusion) by using CNNs. The rationale behind this choice lies in the fact that architectures performing feature fusion have proven beneficial in related problems where combining data from different types of sensors is required \citep{Son2023-pu, Islam2023-sk, Tan2024-lp}. Furthermore, since each signal captures particular properties of the phenomenon of interest using a different sensing principle (sounds of the JM events, and displacement and rotation of the animal head), it is expected that extracting specific features from each of them will be advantageous compared to generating a single signal with multiple channels. 

To solve the problem of JM events recognition (which implies detection and classification), a hybrid multimodal network architecture is presented, composed of multi-head 1D-CNN, RNN, and FNN. To the best of our knowledge, this study represents one of the first multimodal approaches to the problem of JM events recognition using acoustic and IMU signals. The input to the network is represented by frames, which are extracted from the raw signals using fixed sliding time windows without any prior preprocessing or feature extraction. The model classifies each window into one of five possible classes: bite, chew-bite, grazing-chew, rumination-chew, and no-event (to represent the absence of any particular JM event). Hence, the proposed method addresses the challenges of both detecting and classifying JM events simultaneously. 

\begin{figure}
       \centering
       \includegraphics[scale=0.6]{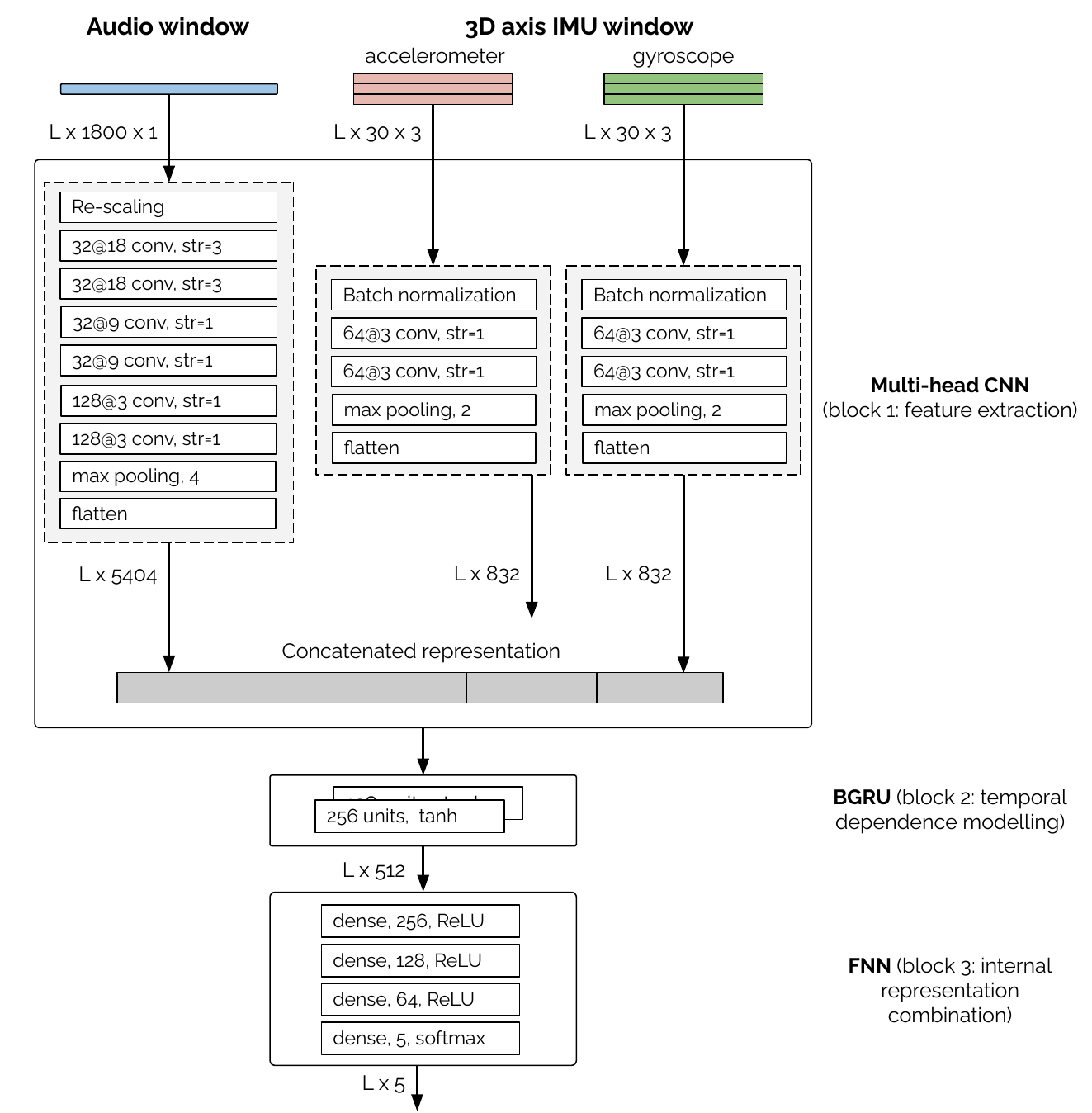}
       \caption{Proposed method architecture: input signals correspond to audio and movement chunks extracted using fixed length time windows. Each convolution layer shows the number of kernels and kernel size (ReLU was used as activiation function), whereas max pooling layers specify the filter size. Dense layers indicate the number of neurons and activation function. At each step the feature dimensions are given, \textcolor{black}{L being} the number of windows in the sequence.}
       \label{fig:proposed_model}
\end{figure}

An overall graphical representation of the proposed model composed of three blocks is presented in Figure~\ref{fig:proposed_model}. The model processes chunks of input signals computed using a time window duration of 300 ms, with a 50\% overlap between consecutive windows. The first block introduces a multi-head CNN combining three independent 1D CNNs. This block extracts low- and high-level features from acoustic and movement signals independently and performs dimensionality reduction at the same time. Each head of the CNN is composed of a normalisation layer (or re-scaling in the audio head), a sequence of 1D convolutional layers, followed by a max pooling layer. A flatten operation is also used in each head, and those values are finally concatenated to create a unique 1D feature vector representation. The second block introduces an RNN, consisting \textcolor{black}{of} a BGRU layer of 256 cells, giving the model the ability to capture temporal dependencies present in data. The last block of the model introduces \textcolor{black}{an} FNN, which combines information in dense layers and predicts class probabilities for each input window. The first and third blocks are enclosed within time-distributed wrappers so that the same layers and parameters are applied to each window of the input sequences. The rectified linear unit (ReLU) was used for all convolutional layers, whilst the cells of the BGRU use hyperbolic tangent and sigmoid. All dense layers of the FNN use ReLU as well, except for the last dense layer, which uses the softmax function for the final classification. The total number of parameters of the model is 11,704,478.

\subsection{Different information fusion strategies}
\label{S:3.3}

As mentioned in Section~\ref{S:1}, there are three main levels at which data fusion can take place: data, features, and decisions. While the proposed model performs feature-level fusion using a multi-head CNN, other architectures that perform fusion at data and decision levels have been proposed and explored as well. 

\begin{figure}
       \centering
       \hspace*{-0.5in}
       \includegraphics[scale=0.5]{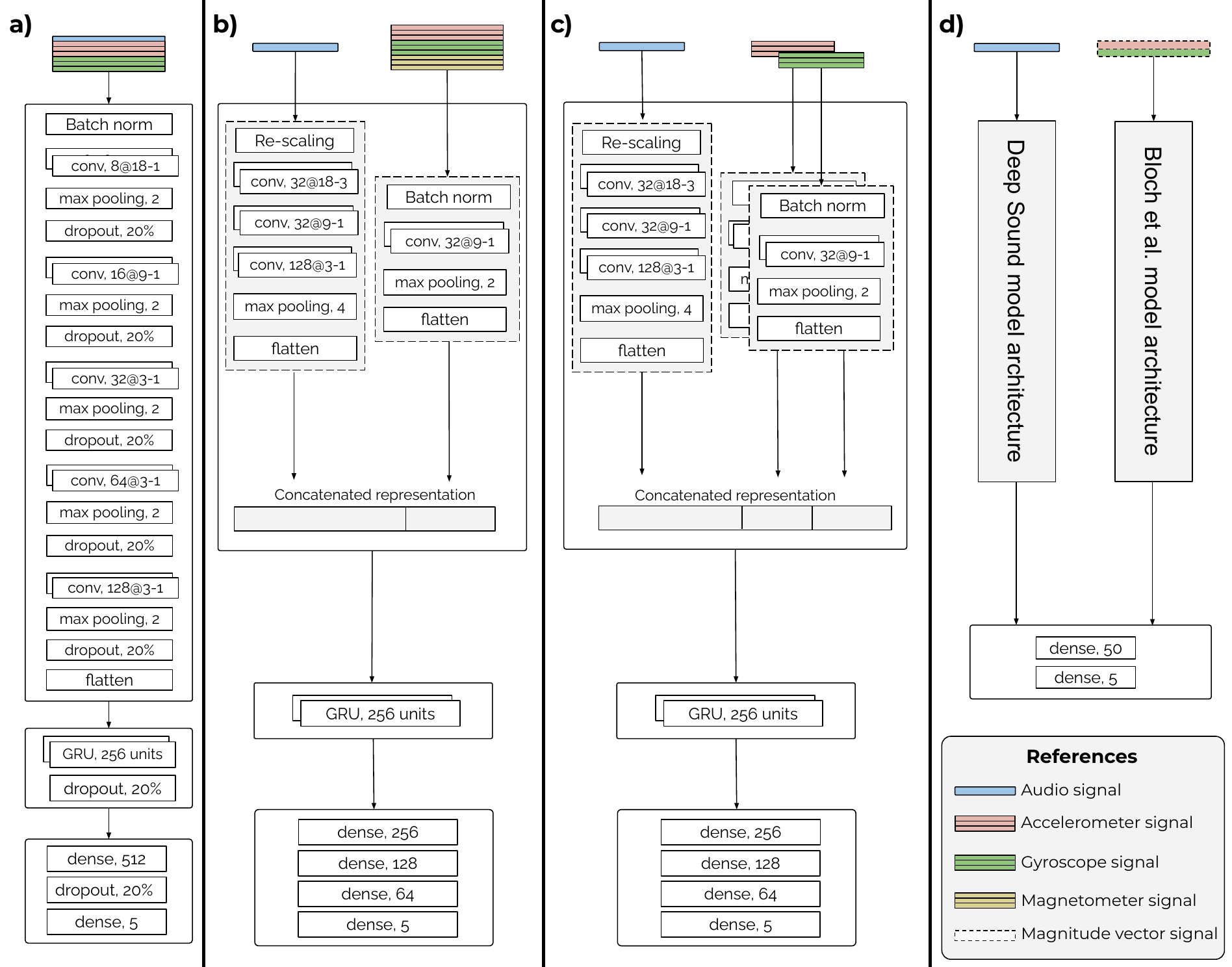}
       \caption{\textcolor{black}{Illustration of the architectures for different fusion levels, where each level represents the configuration that reached the best results.} a) data-level fusion; b) feature fusion with two independent CNN and feature concatenation; c) feature fusion with three independent CNN and feature concatenation (proposed model); d) decision fusion using an FNN for the final decision model. In all cases, the best results were obtained with a window size of 0.3 s.}
       \label{fig:fusion_architectures}
\end{figure}

For comparison purposes, the best-performing model architectures for the different levels of signal fusion were determined in each case (Figure~\ref{fig:fusion_architectures}). In particular, for the feature-fusion level, a variation of the proposed model with 2-heads CNN is included. Several models were evaluated for all fusion levels by varying the number of layers, the size and quantity of filters, and the inclusion of intermediate layers and operations, such as max pooling or dropout (for example, the use of dropout operations has been evaluated in all architectures but it only improves at data-level fusion). Different sizes of the window used to extract data from input signals were also studied. Based on previous studies, durations of 0.3, 0.5, and 1 s were selected for comparison \citep{Alvarenga2020-wc, Ferrero2023-vr}. Different combinations of input signals were also evaluated, using: a) all available raw signals; b) raw audio, accelerometer, and gyroscope signals; c) raw audio signal, and accelerometer and gyroscope vector's magnitude calculated using Eq.\ref{eq:eq7}

\begin{equation}\label{eq:eq7}
s=\sqrt{s_{x}^{2}+s_{y}^{2}+s_{z}^{2}}
\end{equation}

In the data-level fusion architecture (Figure~\ref{fig:fusion_architectures}-a), signals from sound, accelerometer, and gyroscope are concatenated at the initial stage creating a single input to the classifier. Due to differences in the number of samples in each signal, the data from the IMU has been resampled in order to match the sampling frequency of the audio signal. 

Feature-level fusion has been evaluated using a multi-head CNN on two main approaches: i) a 2-head CNN (Figure~\ref{fig:fusion_architectures}-b), which uses one CNN for all data from an IMU sensor; and ii) a 3-head CNN (Figure~\ref{fig:fusion_architectures}-c), which represents the proposed model presented in Section~\ref{S:3.2}. In both cases, an intermediate representation is constructed by doing a concatenation of automatically extracted features from convolutional layers. This combination approach was selected to deal with the difference \textcolor{black}{of} feature space size between heads' outputs, and to provide to the following layers all the available information. Other methods for IMU heads (which share the same input size) were tested - in particular average, maximun, and multiplication - with no improvements.

Decision-level fusion was explored \textcolor{black}{by} implementing two base models which process input signals from each sensor independently (Figure~\ref{fig:fusion_architectures}-d). Audio signals were processed using the architecture proposed by \citet{Ferrero2023-vr}, whereas the proposed architecture by \citet{Bloch2023-xm} was used to process inertial signals. The output probabilities of these models are then introduced to a \textcolor{black}{meta-classifier} to \textcolor{black}{make} a final output decision. Combinations of different base models were also evaluated, including the former two base models, and those proposed by \citet{Chelotti2018-ls} (called Chew-Bite Intelligent Algorithm (CBIA)) and by \citet{Alvarenga2020-wc}. Decision trees and multilayer perceptrons were explored as \textcolor{black}{meta-classifiers}, as well as traditional methods such as majority voting. In all cases, model weights have been initialised randomly.

\subsection{Dataset}
\label{S:3.4}

The fieldwork to collect the dataset occurred on \textcolor{black}{1st} August 2022 at the Campo Experimental J.F. Villarino, Facultad de Ciencias Agrarias, Universidad Nacional de Rosario (UNR) located in the city of Zavalla, Argentina. The area of 450 hectares is made up of several research and productive subsystems, which are representative of the activities in the area of influence (pork, dairy, beef, and crops). In particular, the dairy subsystem can be characterised as a medium-sized, intensified pastoral-based dairy farm with 140-165 milking cows, with an individual daily production of 24-27 l of milk. The protocol used to conduct the experiment has been evaluated and approved by the Committee on Ethical Use of Animals for Research of the UNR. 

The paddock area was approximately \textcolor{black}{1.200 m\textsuperscript{2}} (20 x 60 m) and was fully enclosed with fences. This place was covered with naturalised perennial grasses (with dominance of \textcolor{black}{\textit{Lolium sp}, \textit{Festuca sp} and \textit{Cynodon sp}}). The experimental cows were free to graze within the paddock, and they had permanent access to a watering trough.

\begin{figure}
       \centering
       \includegraphics[scale=0.5]{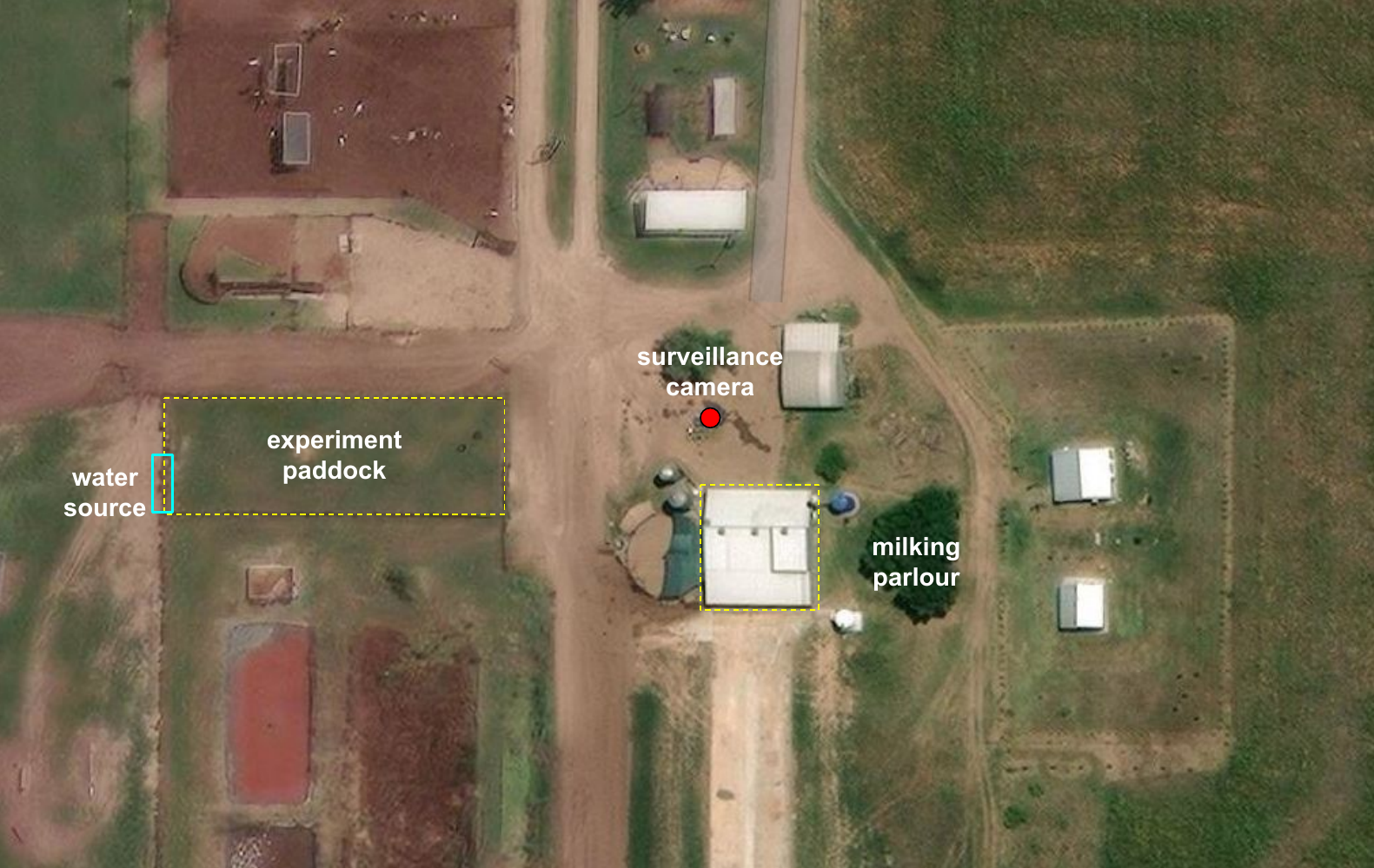}
       \caption{Satellite image of the dairy facilities detailing experimental paddock area, water source, surveillance camera position, and milking parlour.}
       \label{fig:facilities_picture}
\end{figure}

\begin{figure}
       \centering
       \includegraphics[scale=0.50]{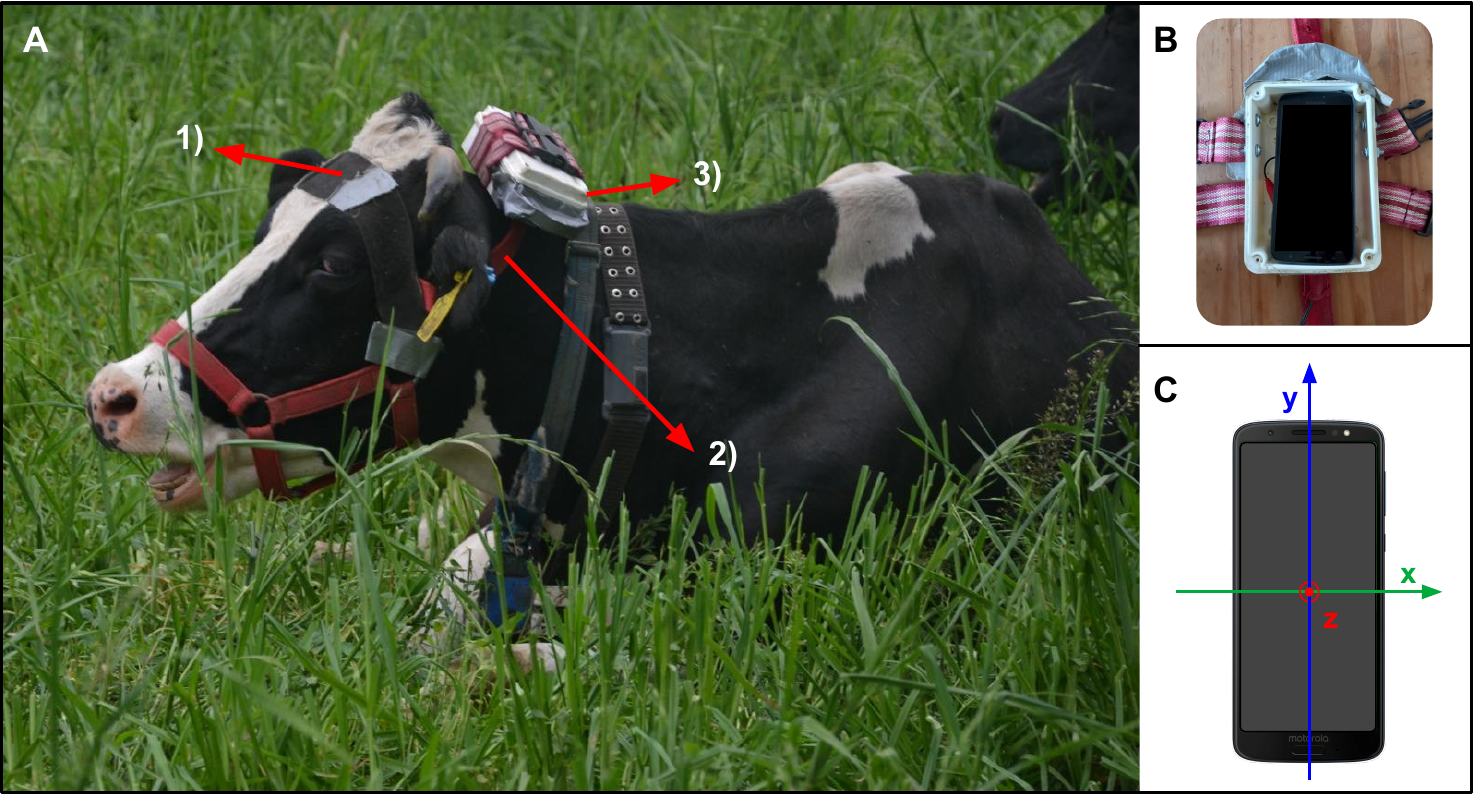}
       \caption{Experimentation setup description. A) Cow in the paddock during a rumination period with external microphone (1), halter (2), and plastic box (3). B) Moto G6 placed in a plastic box; C) axis from IMU sensors orientation: x-axis is aligned with a tail-to-head vector of the animal, y-axis describes sideway movements, whereas z-axis captures up and down movements.}
       \label{fig:recording_setup}
\end{figure}

This area was permanently monitored by an outdoor dome video camera positioned at a lateral distance of 30 \textcolor{black}{meters} from the paddock to assist during the labelling process. Figure~\ref{fig:facilities_picture} introduces a satellite view of the dairy facilities with references to the most important places for the experiment. In addition, two observers with knowledge of animal behaviour manually logged the main behaviours and significant activities on spreadsheets throughout the experiment. Data have been obtained from three 4-year-old lactating Holstein cows weighing 570-600 kg. All cows were tamed and trained in the experimental routine before the final recordings. Each animal was equipped with an acquisition data device consisting of an external microphone (IP57 100 mm, -42 $\pm$ 3 dB, SNR 57 dB) plugged via \textcolor{black}{a} 3.5 mm jack to a Moto G6 smartphone \footnote{Moto G6 smartphone \href{https://www.gsmarena.com/motorola_moto_g6-9000.php}{specifications}.}. Each device was fixed inside a plastic box and secured to prevent internal movements. This same instrumentation has been used in another similar study \citep{Andriamandroso2017-ek}. Microphones were located on the cow’s forehead and covered with rubber foam to isolate them from wind-induced noise and protect them from other frictions. Boxes were mounted to the top side of a halter neck strap (Figure~\ref{fig:recording_setup}).

Data signals were recorded and synchronized using a specifically developed and tested Android application running in the Moto G6 smartphones, using the internal IMU and the external microphones. Three-dimensional IMU signals were recorded using a sampling rate of 100 Hz. Audio recordings were stored using \textcolor{black}{high-efficiency} advanced audio coding \citep{Bosi1997-hb} with a sampling rate of 44.1 kHz and a bit rate of 128 kbps, single channel (mono). The experiment lasted approximately 6 hours (from 09:11:22 to 15:10:20) thus a total of 18 hours were generated in total. For this study, all audio signals were resampled to 6 kHz. Although the experiments were conducted in a confined area, animals were exposed to environmental noise conditions such as \textcolor{black}{bird chirps, wind gusts,} and movements that are not directly related to JM events.

From the collected signals, a total of 29 segments were carefully chosen for annotation with a duration of 9 minutes and 31 seconds on average and a standard deviation (SD) of 1 minute and 57 seconds. Because of the high time demand for the labelling process, a representative subset of signals was selected. A total of 4 hours, 36 minutes and 1.4 seconds have been annotated. The size of the generated dataset represents a significant increase compared to other datasets used in previous studies \citep{Vanrell2020-qa, Martinez-Rau2023-je}. Each segment corresponds to a particular feeding activity (grazing or rumination) and is composed of a sequence of quasi-periodic JM events.

To create \textcolor{black}{event labels}, two experts in ruminant foraging behaviour independently delimited the JM events (including event label, start, and end time) by watching and listening to the acoustic signal. \textcolor{black}{The agreement} result was 97.63\% on average. Both experts worked together to achieve a final decision in case of disagreement.

\begin{figure}
       \centering
       \hspace*{-0.35in}
       \includegraphics[scale=0.55]{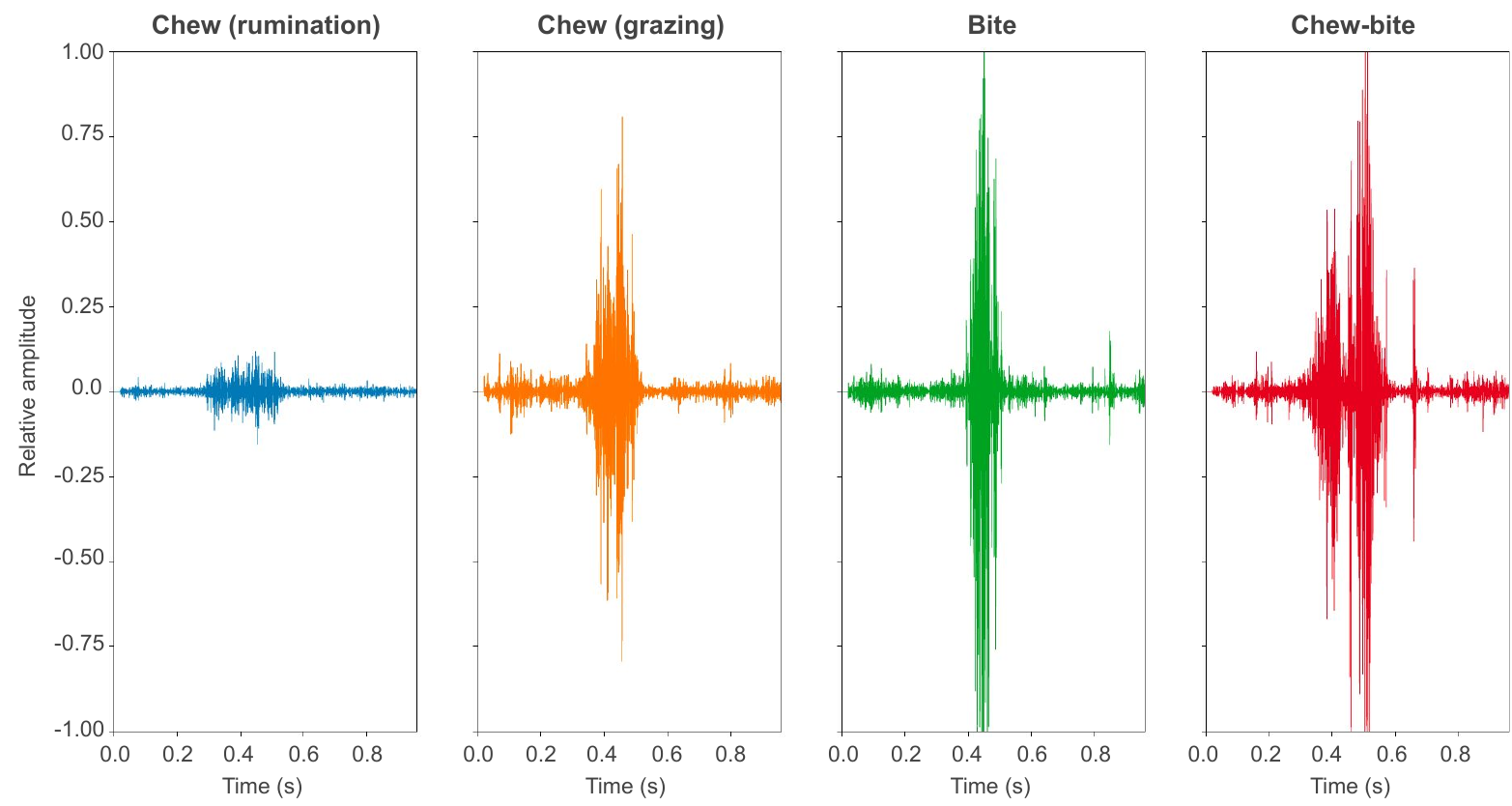}
       \caption{Characteristic waveform of the 4 different JMs events classes considered in the study. Adapted from \citet{Martinez-Rau2023-je}}
       \label{fig:jm_events}
\end{figure}

Based on previous studies \citep{Martinez-Rau2022-ib}, four mutually exclusive labels were treated: bite, grazing-chew, rumination-chew, and chew-bite (a compound movement which is composed of a chew followed by a bite when the animal closes its jaw). Rumination-chew and grazing-chew are events that differ primarily in the feeding activity in which they occur. In the case of rumination, the animal is generally in a state of rest (standing or lying down) and only chew events are present. During grazing, the cow is typically foraging for food (walking, searching, tearing off plants) so the movement of its body and head is recurrent. Chews alternate with bites, or they are even combined (chew-bite). Another difference between rumination-chew and grazing-chew events is the energy of the signal recorded by the acoustic sensor, being higher in the case of grazing \citep{Chelotti2020-pj} \citep{Martinez-Rau2022-jd}. A visual representation of a typical waveform of each JM events from the acoustic signals is presented in Figure~\ref{fig:jm_events}.
The number of labelled samples for each JM event in the dataset and duration statistical values are presented in Table 1.

\begin{table}
	\caption{ Number and duration of annotated jaw movements (JM) events from acoustic signals before windows extraction. }
	\label{tab:tab1}
	\footnotesize
	\centering
	\begin{tabular}{l l c c c}
	\toprule
\multirow{2}{2em}{JM} & \multirow{2}{4em}{Number} & \multicolumn{3}{c}{Duration [s]} \\
& & Mean & Min & Max \\
\midrule

Bite & 2,234 & 0.33 $\pm$ 0.084 & 0.115 & 0.926 \\
Chew-bite & 6,605 & 0.436 $\pm$ 0.087 & 0.187 & 0.961 \\
Grazing-chew & 6,905 & 0.323 $\pm$ 0.066 & 0.144 & 0.665 \\
Rumination-chew & 2,751 & 0.341 $\pm$ 0.051 & 0.167 & 0.806 \\
\textbf{Overall} & 18,495 & 0.362 $\pm$ 0.092 & 0.115 & 0.961 \\
\bottomrule
	\end{tabular}
\end{table}

\section{Experiments, results and discussions}
\label{S:5}

In this section, the methodology selected to drive the experimentation is explained, and the results and discussions of performed experiments are presented as well.

\subsection{Experimental settings}
\label{S:5.1}

From the total of 29 signal segments, 24 were used for model selection purposes. All models were trained and evaluated using a 5-fold \textcolor{black}{cross-validation} (CV) scheme with each fold containing 4 or 5 segments. Each fold contains 1 segment from a rumination period and the rest from grazing intervals. This relation between grazing and rumination was proposed to balance the number of JM events. While grazing includes grazing-chews, bites and chew-bites, rumination only contains ruminating-chews. The remaining 5 segments were separated for test purposes, meaning the evaluation of the generalization capability of the model performing the best on validation sets. The separation of data into different sets was conducted before the experimentation stage, and these sets remained constant throughout this stage. In order to solve class imbalance, the weights of training samples were adapted according to Eq.~\ref{eq:eq8}.

\begin{equation}\label{eq:eq8}
W_{ic}=\frac{N_{max}}{N_c}
\end{equation}

where $W_{ic}$ is the weight of instance $i$ associated with class $c$; $N_{max}$ is the number of instances of the majority class and $N_c$ is the number of instances of class $c$. Experiments using a data augmentation approach \textcolor{black}{were evaluated} as an alternative to sample weighting to solve the class imbalance, with inferior results.

For unification process during training, all windows extracted from each signal were converted into smaller sequences of a fix number of windows. Based on this, each example provided to the model consists of a sequence of L windows. Different values have been evaluated for this parameter and L=46 emerged as the one that obtained the best results in preliminary experiments. The length of the original signal included in each sequence varies according to the window size, being for example 6.9 seconds for a window size of 300 ms with 50\% overlap. A padding operation was used to complete the missing windows in those shorter sequences if necessary.

All the necessary code was developed using Python version 3.10.12 and it is available in the project repository \citep{Ferrero2024}. Several utilities from Python library scikit-learn 1.2.2 have been used, in particular label encoders, k-fold extraction, grid search, and the implementation of traditional machine learning algorithms (such as decision trees). Tensorflow 2.12.0 was used to define and train the neural \textcolor{black}{network} architectures. Experiments were performed using an Intel Core™ i7-8700 3.20GHz CPU, 64 GB RAM and a dual NVIDIA GPU configuration composed of 24 GB GeForce RTX 3090 and 24 GB RTX A5000.

For training, the Adam optimiser \citep{Kingma2014-yi} was chosen, utilising a total of 1400 epochs with an early stopping tolerance of 50 epochs. The batch size was set to 5, and categorical cross-entropy was employed as the loss function. Default values were retained for the remaining parameters.

\subsection{Evaluation metrics}
\label{S:5.2}

The process of JM events recognition involves initially detecting the event, i.e., recognizing the onset and offset, and subsequently, assigning a class to the event. In this scenario, detection errors directly impact the classification task. Based on this, the problem addressed in this work requires the use of an evaluation methodology that takes into account both aspects.

The \textit{sed\_eval} toolbox \citep{Mesaros2016-xe, Mesaros2021-sv} has been selected to calculate the performance during experimentation. This tool has been used in numerous studies related to event recognition in sounds \citep{Serizel2020-qc, Venkatesh2022-xd, Ferrero2023-vr}. Furthermore, it is a comprehensive open-source toolbox that implements a range of metrics suitable for the objectives set in this work.

\begin{figure}
       \centering
       \includegraphics[scale=0.75]{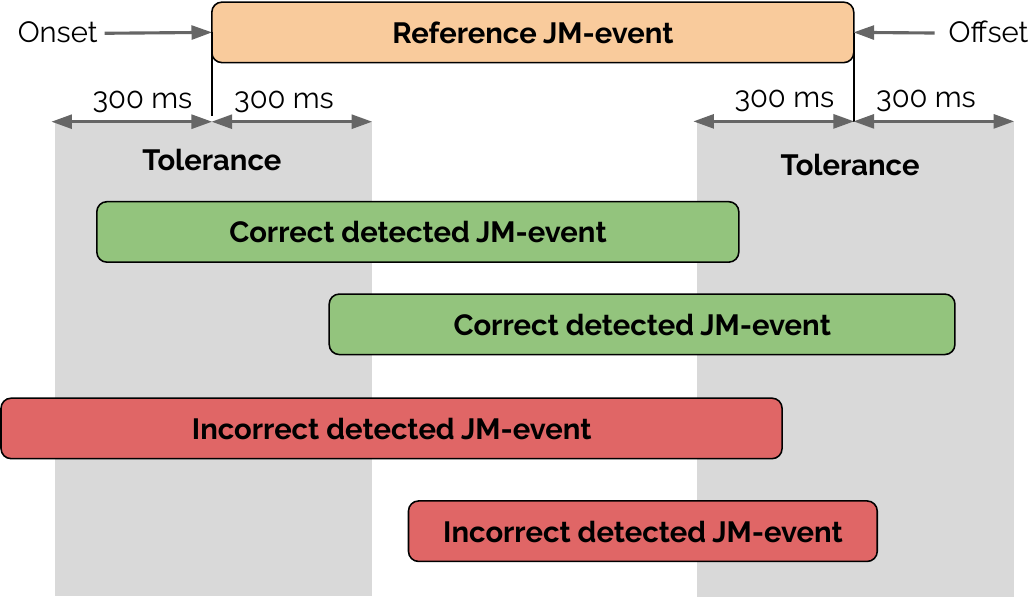}
       \caption{Illustration of evaluation procedure implemented by the sed\_eval toolbox used in this article. Two pairs of JM events (one pair correct and one pair incorrect) with respect to a reference JM event. Adapted from \citet{Mesaros2016-xe, Mesaros2021-sv}.}
       \label{fig:evaluation_procedure}
\end{figure}

Given a reference event, the criterion used by the tool to \textcolor{black}{classify} a prediction generated by a system as correct includes three conditions: \textbf{a)} the onset of the predicted event must fall within the interval defined by the onset of the reference event $\pm$ tolerance value (300 ms); \textbf{b)} the offset of the predicted event must fall within the interval defined by the offset of the reference event $\pm$ tolerance value (300 ms); \textbf{c)} the class of both events must be equivalent. Figure~\ref{fig:evaluation_procedure} introduces examples where different situations for conditions a) and b) can be observed.

Regarding classification results, the metrics expressed in Eq.~\ref{eq:eq9} to \ref{eq:eq12} have been used:

\begin{equation}\label{eq:eq9}
precision = \frac{TP}{TP + FP}
\end{equation}

\begin{equation}\label{eq:eq10}
recall = \frac{TP}{TP + FN}
\end{equation}

\begin{equation}\label{eq:eq11}
F1-score = \frac{2 \cdot precision \cdot recall}{precision + recall }
\end{equation}

\begin{equation}\label{eq:eq12}
Error-rate = \frac{S + D + I}{N}
\end{equation}

where $TP$ denotes true positive, $FP$ false positive, $FN$ false negative, $S$ substitutions (correct detected JM events in system output but incorrectly labelled), $I$ insertions (detected JM events for the system output that do not exist in the ground truth) and $D$ deletions (ground truth JM events that are not detected). Metrics were computed for each class individually as well as for the overall multi-class. The overall metrics handle the multi-class imbalanced condition by computing micro (class) averages \citep{Sokolova2009-tx}. Micro average computation implies that $TP$, $FP$, and $FN$ are obtained by summing up samples through all classes. For instance, the term $TP$ is ultimately represented as $TP_{gc}$ + $TP_{rc}$ + $TP_{cb}$ + $TP_b$, denoting the number of true positives for grazing-chews, rumination-chews, chew-bites, and bites, respectively. With the exception of the error rate, for all other metrics between Eq. \ref{eq:eq9} to \ref{eq:eq12} \textcolor{black}{higher values are indicative of a better model}.

\subsection{Fusion level comparison}
\label{S:5.3}

An evaluation of the classification performance of the considered level fusion architectures from Figure~\ref{fig:fusion_architectures} is presented in Table 2. The information fusion scheme that achieved the best results was the feature-level in all analysed metrics. In particular, the proposed model with 3-heads CNN scored the best based on the overall F1-score. In addition, the decision-level model outperformed the data-level architecture. For all metrics, data-level fusion presented the lowest performance. Particularly remarkable is the incapacity of this architecture to recognise associated rumination-chews events.

One possible interpretation of this comparison might be that feature-level fusion is more suitable for this task compared to data-level and decision-level fusion because it leverages the strengths of both sensor modalities (acoustic and inertial signals) by integrating informative features that are automatically extracted from each sensor domain. This might be interpreted as a specific \textcolor{black}{characteristic} of the presented approach, due to different conclusions reported in \textcolor{black}{other} studies with other model specifications \citep{Nweke2019-dr, Arablouei2023-lg}. In data-level fusion, raw sensor data from different modalities are directly combined, which can lead to issues due to the heterogeneity of the data types (e.g., sampling rates, signal formats), making it difficult for the model to effectively exploit the complementary nature of each sensor. On the other hand, decision-level fusion combines predictions from separate models trained on individual sensor types. While this approach might capture some modality-specific insights, it does not capitalize on the synergistic relationships between features from different sensors during the learning process.

Although the structures of both feature-fusion level architectures are similar, the 2-head CNN architecture obtained slightly inferior results. Apart from that, the use of magnetometer signals does not appear to provide benefits over using only the accelerometer and gyroscope signals. This is probably related to the fact that the execution of JM events does not have a relation with changes to any particular location, something that is measured by this sensor. In fact, the performance of the model drops when using this signal, due to the need to process data that apparently does not contain discriminative power in this context.

When comparing the recognition performance for individual JM event classes, worse results were obtained with the minority class (bite), even with the use of different weights per class to counteract the data imbalance. These results are consistent with previous studies \citep{Martinez-Rau2022-ib, Ferrero2023-vr}. 

\begin{figure}
       \centering
        \hspace*{-0.35in}
       \includegraphics[scale=0.1]{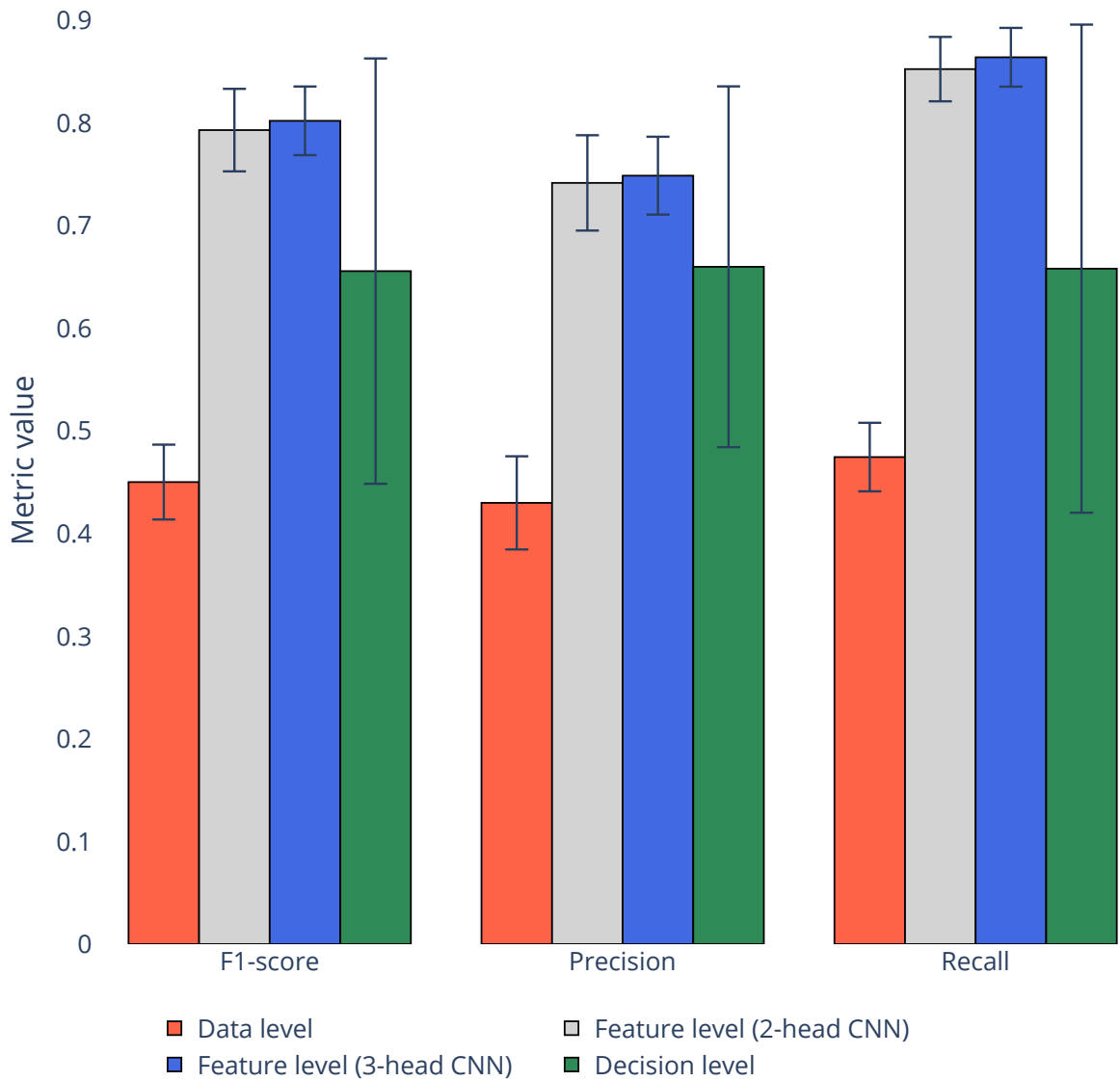}
       \caption{Comparison of the results obtained by different fusion levels based on the \textcolor{black}{overall} F1-score, precision, and recall.}
       \label{fig:fusion_results}
\end{figure}

Overall, there is a reduced variability in the metrics (F1-score, precision, and recall) obtained across the first three fusion levels, indicating stability in the performance of the models (Figure~\ref{fig:fusion_results}). The decision-level model is an exception because \textcolor{black}{the} SD between the folds is significant.

\begin{table}
        \setlength\extrarowheight{-3pt}
	\caption{Information fusion architectures (Figure~\ref{fig:fusion_architectures}) results based on F1-score, precision, recall, and error rate by class and overall results. In all cases, the average and the SD across validation sets during the 5-fold CV phase are reported. }
	\label{tab:tab2}
	\footnotesize
	\centering
 	\resizebox{\textwidth}{!}{%
    	\begin{tabular}{l c c c c}
    	\toprule
\multirow{2}{2em}{} & Data level & Feature level & Feature level & Decision level\\
\multirow{2}{2em}{} &  & (2-heads CNN) & (3-heads CNN) & \\
\midrule
& \multicolumn{4}{c}{\textbf{F1-score}} \\
\midrule
Bite & 0.403 $\pm$ 0.066 & 0.581 $\pm$ 0.096 & \textbf{0.662 $\pm$ 0.006} & 0.469 $\pm$ 0.274 \\
Chew-bite & 0.624 $\pm$ 0.035 & 0.797 $\pm$ 0.005 & \textbf{0.811 $\pm$ 0.027} & 0.733 $\pm$ 0.145 \\
Grazing-chew & 0.389 $\pm$ 0.041 & \textbf{0.809 $\pm$ 0.026} & 0.805 $\pm$ 0.038 & 0.562 $\pm$ 0.334 \\
Rumination-chew & 0.013 $\pm$ 0.022 & \textbf{0.870 $\pm$ 0.049} & 0.827 $\pm$ 0.146 & 0.670 $\pm$ 0.195 \\
\textbf{Overall} & 0.450 $\pm$ 0.036 & 0.793 $\pm$ 0.040 & \textbf{0.802 $\pm$ 0.033} & 0.656 $\pm$ 0.207 \\
\midrule
& \multicolumn{4}{c}{\textbf{Precision}} \\
\midrule
Bite & 0.357 $\pm$ 0.134 & \textbf{0.758 $\pm$ 0.051} & 0.717 $\pm$ 0.039 & 0.587 $\pm$ 0.147 \\
Chew-bite & 0.517 $\pm$ 0.033 & 0.717 $\pm$ 0.084 & \textbf{0.747 $\pm$ 0.052} & 0.663 $\pm$ 0.183 \\
Grazing-chew & 0.386 $\pm$ 0.052 & \textbf{0.728 $\pm$ 0.025} & 0.719 $\pm$ 0.050 & 0.656 $\pm$ 0.186 \\
Rumination-chew & 0.062 $\pm$ 0.085 & 0.856 $\pm$ 0.026 & \textbf{0.866 $\pm$ 0.029} & 0.676 $\pm$ 0.192 \\
\textbf{Overall} & 0.430 $\pm$ 0.045 & 0.742 $\pm$ 0.046 & \textbf{0.749 $\pm$ 0.038} & 0.660 $\pm$ 0.176 \\
\midrule
& \multicolumn{4}{c}{\textbf{Recall}} \\
\midrule
Bite & 0.528 $\pm$ 0.090 & 0.488 $\pm$ 0.130 & \textbf{0.618 $\pm$ 0.081} & 0.445 $\pm$ 0.297 \\
Chew-bite & 0.788 $\pm$ 0.058 & \textbf{0.908 $\pm$ 0.022} & 0.890 $\pm$ 0.010 & 0.839 $\pm$ 0.074 \\
Grazing-chew & 0.397 $\pm$ 0.046 & 0.910 $\pm$ 0.028 & \textbf{0.917 $\pm$ 0.018} & 0.559 $\pm$ 0.377 \\
Rumination-chew & 0.007 $\pm$ 0.013 & \textbf{0.887 $\pm$ 0.081} & 0.822 $\pm$ 0.216 & 0.674 $\pm$ 0.202 \\
\textbf{Overall} & 0.474 $\pm$ 0.033 & 0.852 $\pm$ 0.031 & \textbf{0.864 $\pm$ 0.029} & 0.658 $\pm$ 0.238 \\
\midrule
& \multicolumn{4}{c}{\textbf{Error rate}} \\
\midrule
Bite & 1.643 $\pm$ 0.504  & 0.674 $\pm$ 0.084 & \textbf{0.624 $\pm$ 0.090} & 0.786 $\pm$ 0.221 \\
Chew-bite & 0.950 $\pm$ 0.094 & 0.471 $\pm$ 0.149 & \textbf{0.418 $\pm$ 0.077} &  0.669 $\pm$ 0.437 \\
Grazing-chew & 1.248 $\pm$ 0.133 & \textbf{0.431 $\pm$ 0.058} & 0.447 $\pm$ 0.101 & 0.625 $\pm$ 0.340 \\
Rumination-chew & 1.053 $\pm$ 0.045 & \textbf{0.262 $\pm$ 0.086} & 0.302 $\pm$ 0.197 & 0.662 $\pm$ 0.401 \\
\textbf{Overall} & 1.015 $\pm$ 0.107 & 0.337 $\pm$ 0.064 & \textbf{0.327 $\pm$ 0.050} & 0.513 $\pm$ 0.31 \\
\bottomrule
	\end{tabular}
 }
\end{table}

\subsection{Effect of time window size and quantization}
\label{S:5.4}

The performance of the proposed model has been evaluated for different sliding input window sizes. Table~\ref{tab:tab3} introduces the results using three different sizes of sliding windows: 0.3 s, 0.5 s, and 1 s. The overlap between two consecutive windows was 50\%. The reported results include the average values per metric for the different validation folds, as well as the SD.

A window size of 0.3 s exhibited the best metrics, while the use of 1 s windows performed the worst. Based on these results, it would be useful to use short time windows, similar to the average duration of JM events (Table~\ref{tab:tab1}).

\begin{figure}
       \centering
       \includegraphics[scale=0.35]{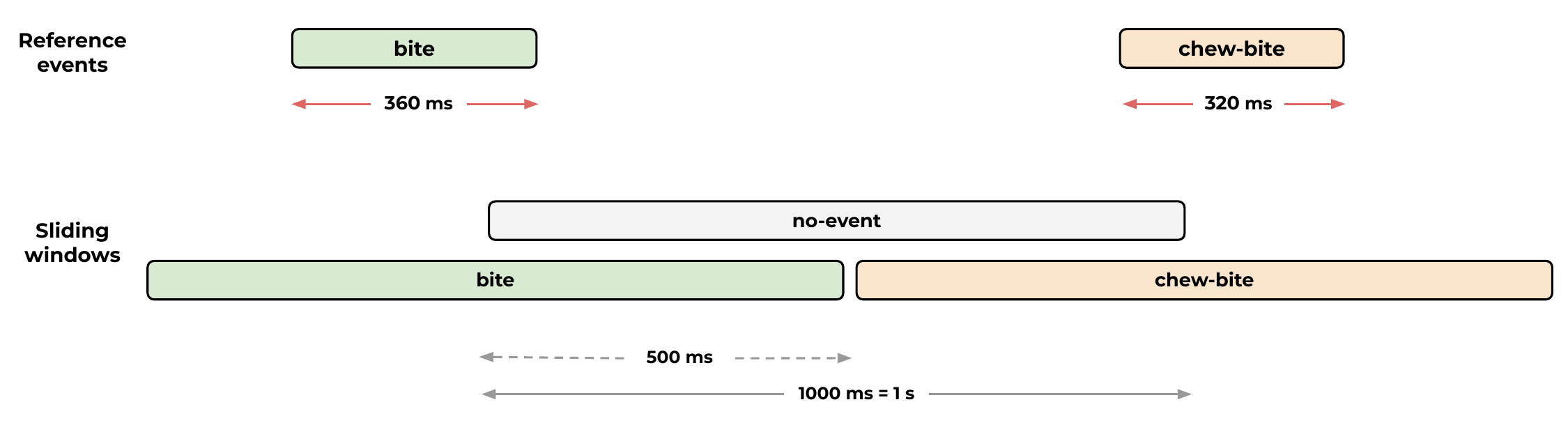}
       \caption{An example representation for a time window length of 1000 ms = 1-s with two reference events and three extracted windows.}
       \label{fig:time_windows_effect}
\end{figure}

Conversely, the use of longer time windows seems to worsen the performance. There are two likely causes for this: firstly, when extracting 1 s fragments, two consecutive JM events could be partially included, generating chunks with valuable information that are categorised as \textcolor{black}{the} absence of JM events - "no-event" class (Figure~\ref{fig:time_windows_effect}). Lastly, the detection of JM events represents a challenge for the tolerance value selected for evaluation purposes, since JM events generally have a duration shorter than the window size.

\begin{table}
        \setlength\extrarowheight{-3pt}
	\caption{ Performance of the proposed model with 0.3, 0.5, and 1 s time windows, each with a 50\% overlap. }
	\label{tab:tab3}
	\footnotesize
	\centering
 	\resizebox{\textwidth}{!}{%
    	\begin{tabular}{l c c c c}
    	\toprule
\multirow{2}{2em}{} & F1-score & Precision & Recall & Error rate \\
\midrule
0.3 s & \textbf{0.802 $\pm$ 0.033} & \textbf{0.749 $\pm$ 0.038} & \textbf{0.864 $\pm$ 0.029} & \textbf{0.327 $\pm$ 0.05} \\
0.5 s & 0.507 $\pm$ 0.254 & 0.496 $\pm$ 0.238 & 0.524 $\pm$ 0.263 & 0.769 $\pm$ 0.245 \\
1 s & 0.297 $\pm$ 0.229 & 0.314 $\pm$ 0.207 & 0.295 $\pm$ 0.233 & 1.006 $\pm$ 0.119 \\
\bottomrule
	\end{tabular}
 }
\end{table}

\begin{table}
        \setlength\extrarowheight{-3pt}
	\caption{ Comparison of presented model results using quantization. }
	\label{tab:tab4}
	\footnotesize
	\centering
 	\resizebox{\textwidth}{!}{%
    	\begin{tabular}{c c c c c}
    	\toprule
Weights precision & F1-score & Precision & Recall & Error rate \\
\midrule
float 32 & \textbf{0.802 $\pm$ 0.033} & \textbf{0.749 $\pm$ 0.038} & \textbf{0.864 $\pm$ 0.029} & \textbf{0.327 $\pm$ 0.05} \\
float 16 & 0.791 $\pm$ 0.05 & 0.742 $\pm$ 0.054 & 0.848 $\pm$ 0.045 & 0.335 $\pm$ 0.069 \\
\bottomrule
	\end{tabular}
 }
\end{table}

Quantization is employed to optimize the model by lowering the precision of weights, which reduces memory usage and computational requirements, making the model more suitable for deployment on resource-constrained devices. Results from the exploration of post-training quantization are presented in Table~\ref{tab:tab4}, which helps to analyze their effects on key metrics such as F1-score, precision, recall, and error rate. Those results demonstrate that quantization using a float 16 precision (instead of the default float 32) achieves significant efficiency gains while maintaining acceptable performance levels.

\subsection{Comparison between the proposed model and state-of-the-art methods}
\label{S:5.5}

The performance of the proposed model (Figure~\ref{fig:fusion_architectures}-c) was compared against different state-of-the-art methods. Four unimodal models were selected to encompass four combinations, integrating audio and movement signals using both traditional and deep neural network methods:

\begin{enumerate}
    \item The CBIA is a pattern recognition method that processes acoustic signals to perform event detection using thresholds, feature extraction over the detected event, and then classification using an FNN \citep{Chelotti2018-ls}.
    \item The Deep Sound architecture combines convolutional, recurrent, and fully connected layers to recognise (detect and classify) JM events using sliding windows \citep{Ferrero2023-vr}.
    \item The traditional approach proposed by \citet{Alvarenga2020-wc} processes motion signals using sliding windows, and a specific feature engineering process is proposed for the classification of short-duration activities in ruminants for each window.
    \item The deep architecture proposed by \citet{Bloch2023-xm} consists of CNN and FNN to recognise feeding activities in ruminants using motion input signals.
\end{enumerate}

All selected methods have been trained and validated using the same dataset partitions that were used for the exploration of the fusion level architectures.

The average and SD values for the different validation partitions during the 5-fold CV process are shown in Table~\ref{tab:tab5}. It can be seen that for all analysed metrics, the proposed model outperforms all unimodal methods, while the Deep Sound and CBIA models are in second and third performance rank, respectively. Regarding the unimodal approaches, there is \textcolor{black}{a} remarkable improvement in acoustic methods (CBIA and Deep Sound) compared to movement-based methods (Alvarenga and Bloch). This acknowledges the previous statement regarding the advantages of sound over inertial signals to recognise JM events. On the other hand, even though the use of deep architectures offers better results in sound processing, the opposite occurs in the case of signals extracted from the IMU.

Different input signal alternatives were evaluated for the model proposed by \citet{Bloch2023-xm}, including the use of raw signals, the calculation of magnitude vectors from each signal, and the use of band-pass filters (as described by the authors) as well as their omission. The results obtained in all cases were worse than those reported in Table~\ref{tab:tab5} (where the Hamming filter proposed by the authors was included and all the raw signals were used as input). 

As previously mentioned, for movement-signals-based options is it noteworthy that the deep learning models underperform the classic models. This suggests, in conjunction with the results reported by \citet{Alvarenga2020-wc}, that a more exhaustive exploration of deep architectures that allow automatically obtaining more representative variables from data could be beneficial.

On the other hand, the results from motion methods are observed to be significantly lower. This seems to indicate a clear difficulty in recognising short-duration events (such as JM events) using these signals, which is aligned with what was previously mentioned in the related work section. It can be established that this type of \textcolor{black}{signal offers} an advantage when used in conjunction with audio signals, but their independent use in this problem is insufficient.

\begin{table}
        \setlength\extrarowheight{-3pt}
	\caption{ Overall results obtained for the multi-head CNN-RNN fusion proposed model and selected state-of-the-art algorithms. }
	\label{tab:tab5}
	\footnotesize
	\centering
 	\resizebox{\textwidth}{!}{%
    	\begin{tabular}{l c c c c}
    	\toprule
\multirow{2}{2em}{} & F1-score & Precision & Recall & Error rate \\
\midrule
Alvarenga et al. (2020) & 0.251 $\pm$ 0.015 & 0.188 $\pm$ 0.015 & 0.381 $\pm$ 0.008 & 1.977 $\pm$ 0.129 \\
Bloch et al. (2023) & 0.125 $\pm$ 0.009 & 0.123 $\pm$ 0.012 & 0.127 $\pm$ 0.007 & 1.615 $\pm$ 0.067 \\
CBIA & 0.606 $\pm$ 0.066 & 0.627 $\pm$ 0.063 & 0.587 $\pm$ 0.072 & 0.499 $\pm$ 0.074 \\
Deep Sound & 0.704 $\pm$ 0.025 & 0.650 $\pm$ 0.030 & 0.767 $\pm$ 0.020 & 0.453 $\pm$ 0.052 \\
Proposed model & \textbf{0.802 $\pm$ 0.033} & \textbf{0.749 $\pm$ 0.038} & \textbf{0.864 $\pm$ 0.029} & \textbf{0.327 $\pm$ 0.05} \\
\bottomrule
	\end{tabular}
 }
\end{table}

\subsection{Ablation study and test performance}
\label{S:5.6}

\begin{figure}
       \centering
       \hspace*{-0.5in}
       \includegraphics[scale=0.45]{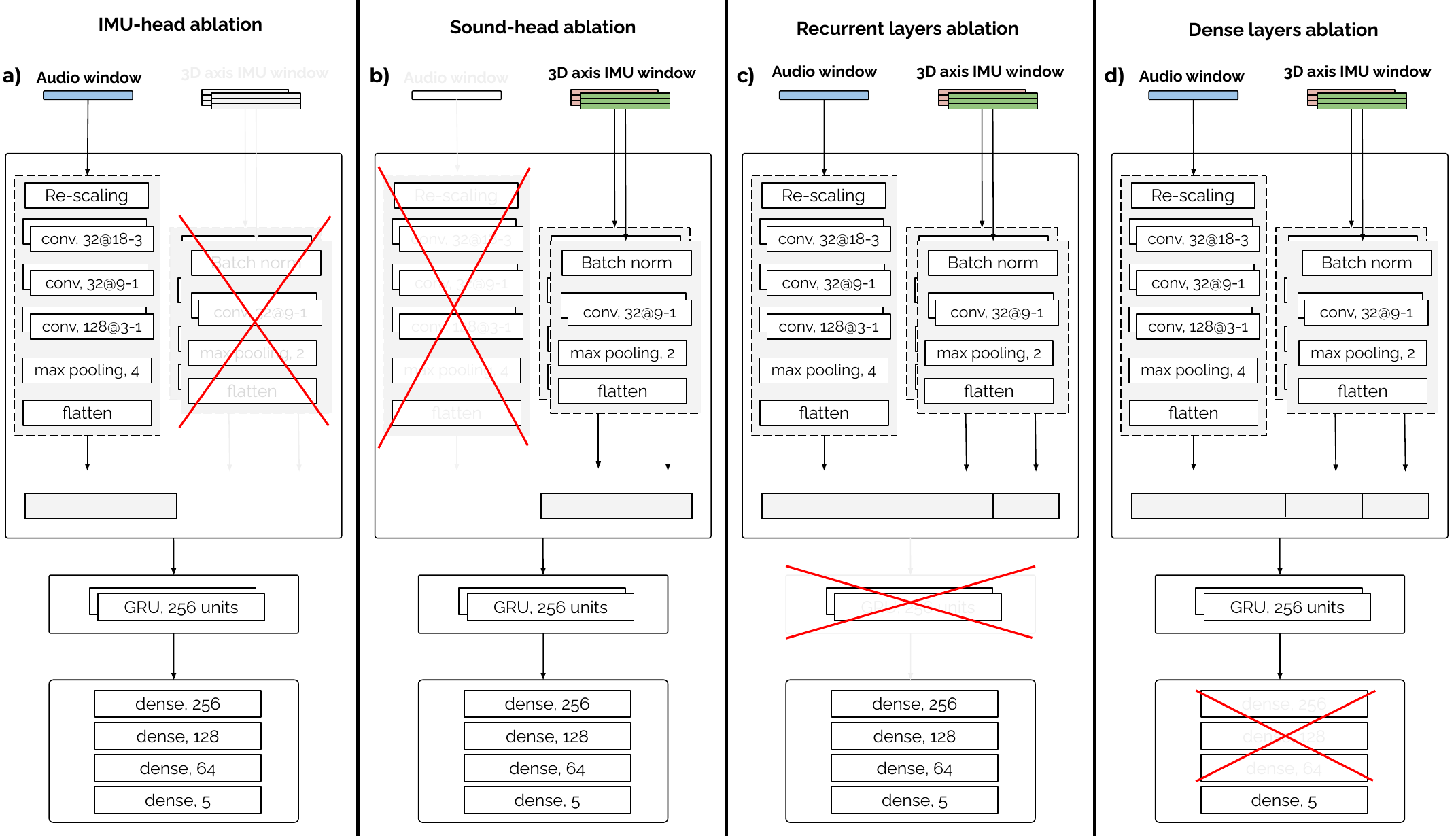}
       \caption{Different architectures proposed in the ablation study. a) proposed model with only sound head; b) proposed model with only IMU head; c) proposed model with no recurrent block; d) proposed model with only one dense layer in the last block.}
       \label{fig:ablation_study}
\end{figure}

In order to evaluate the capabilities of each component in the proposed model, four different ablation experiments have been conducted. The architectures explored in those experiments are introduced in Figure~\ref{fig:ablation_study} highlighting the differences with the proposed model. Two of them were focused on the input blocks: Figure~\ref{fig:ablation_study} a) the proposed model without IMU heads and Figure~\ref{fig:ablation_study} b) the proposed model without \textcolor{black}{the} sound head. These experiments are important from a practical point of view. During the execution of a multimodal system, if one of the inputs is lost or has strong interference, the performance could be severely affected. In this situation, it is often convenient to discard one of the inputs. In the context of this application specifically, this is commonly seen in environments where animals are confined (barn). In these cases, the signal-to-noise ratio of the sound is low due to the noise and reverberations and it is often convenient to discard this data and use only motion data. The execution of experiments with controlled noise in future studies will allow an evaluation \textcolor{black}{of} which signal is most convenient to be used in these scenarios, noisy sound or IMU signals.

The remaining two experiments were focused on specific blocks of the original model: Figure~\ref{fig:ablation_study} c) the proposed model without recurrent layers (block 2) and Figure~\ref{fig:ablation_study} d) the proposed model with only the last dense layer in block 3. These experiments seek to simplify the structure of the proposed model without greatly affecting performance. Simplifying the model can reduce the risk of overfitting and the amount of data needed for the model to achieve good performance.

\begin{table}
        \setlength\extrarowheight{-3pt}
	\caption{Performance of the ablation study including four architectures and the proposed model for cross-validation folds and test set. Inference time refers to calculations to process 1 minute of signal. V: Validation. T: Test. PM: Proposed model. b.1 and b.2 reflect the extraction of gyroscope and accelerometer head from the architecture proposed in b).}
	\label{tab:tab6}
	\footnotesize
	\centering
 	\begin{adjustbox}{max width=\textwidth}
    	\begin{tabular}{l c c c c c c c c}
    	\toprule
\multicolumn{2}{c}{} & F1-score & Precision & Recall & Error rate & Parameters & FLOPs & Inference time (s) \\
\toprule
\multirow{2}{2em}{a} & V & 0.576 & 0.542 & 0.615 & 0.536 & \multirow{2}{*}{11,678,470} & \multirow{2}{*}{$9.3\times 10^{10}$} & \multirow{2}{*}{0.215 $\pm$ 0.031} \\
& T & 0.686 & 0.660 & 0.713 & 0.388 & &  \\

  \arrayrulecolor{black!30}
  \midrule
  \arrayrulecolor{black}

\multirow{2}{2em}{b} & V & 0.155 & 0.182 & 0.156 & 1.144
 & \multirow{2}{*}{11,605,214} & \multirow{2}{*}{$8.6\times 10^{9}$} & \multirow{2}{*}{0.211 $\pm$ 0.014} \\
& T & 0.001 & 0.087 & 0.001 & 1.004 & &  \\

  \arrayrulecolor{black!30}
  \midrule
  \arrayrulecolor{black}

\multirow{2}{2em}{b.1} & V & 0.011 & 0.068 & 0.006 & 1.044
 & \multirow{2}{*}{11,605,214} & \multirow{2}{*}{$8.5\times 10^{9}$} & \multirow{2}{*}{0.208 $\pm$ 0.008} \\
& T & 0.001 & 0.026 & 0.001 & 1.016 & &  \\

  \arrayrulecolor{black!30}
  \midrule
  \arrayrulecolor{black}

\multirow{2}{2em}{b.2} & V & 0.165 & 0.185 & 0.167 & 1.150
 & \multirow{2}{*}{11,605,214} & \multirow{2}{*}{$8.5\times 10^{9}$} & \multirow{2}{*}{0.208 $\pm$ 0.008} \\
& T & 0.002 & 0.033 & 0.001 & 1.023 & &  \\

  \arrayrulecolor{black!30}
  \midrule
  \arrayrulecolor{black}

\multirow{2}{2em}{c} & V & 0.607 & 0.473 & 0.851 & 1.008 & \multirow{2}{*}{298,142} & \multirow{2}{*}{$3.7\times 10^{7}$} & \multirow{2}{*}{0.133 $\pm$ 0.008} \\
& T & 0.574 & 0.437 & 0.838 & 1.146 & &  \\

  \arrayrulecolor{black!30}
  \midrule
  \arrayrulecolor{black}

\multirow{2}{2em}{d} & V & 0.738 & 0.690 & 0.795 & 0.427 & \multirow{2}{*}{11,531,998} & \multirow{2}{*}{$1.59\times 10^{11}$} & \multirow{2}{*}{0.209 $\pm$ 0.009} \\
& T & 0.743 & 0.697 &0.795 & 0.444 & &  \\

  \arrayrulecolor{black!30}
  \midrule
  \arrayrulecolor{black}

\multirow{2}{2em}{PM} & V & 0.802 & 0.749 & 0.861 & 0.325 & \multirow{2}{*}{11,704,478} & \multirow{2}{*}{$1.59\times 10^{11}$} & \multirow{2}{*}{0.217 $\pm$ 0.034} \\
& T & \textbf{0.813} & \textbf{0.771} & \textbf{0.859} & \textbf{0.306} & &  \\
\bottomrule
	\end{tabular}
\end{adjustbox}
\end{table}

The results of the ablation study in terms of performance metrics, number of model parameters, floating point operations (FLOPs), and inference time are presented in Table~\ref{tab:tab6}, including the average performance on validation folds as well as on the test set. It can be observed that in all cases, the elimination of a specific part from the proposed model worsens the performance, pointing out that all parts play an important role and have an impact on the final architecture.

The worst results were exhibited by option b), that is, using only motion data. These results were expected because, in the particular case of JM events recognition, sound signals offer more discriminative power than motion signals \citep{Chelotti2023-gk}. This option also shows convergence issues when trying to predict the test set. Furthermore, the concept of option a) (considering only the sound input) achieves similar results in the test set to those indicated in the CNN-RNN acoustic method \citep{Ferrero2023-vr}. When using only motion data, \textcolor{black}{ the  gyroscope head (b.2) performs better in general than accelerometer head (b.1) both on validation and test data}.

The final dense layers of the FNN are responsible for generating the final output of the model by combining the features obtained in the previous layers. Removing this set of layers reduces the overall performance of the model, as can be seen in results achieved by option d). This option achieved the best performance of the four ablated models (except for recall where option c) reported the higher values), but still underperformed the proposed model.
Moreover, given the large number of parameters of the proposed model, the removal of all recurrent layers simplifies the model and considerably reduces the risk of overfitting. When removing these layers - option d) -, the model performance was also highly damaged, thus confirming the importance of the temporal component in this problem. To the best of our knowledge, no specific study confirms the temporal dependence of this phenomenon. However, there are works based on Hidden Markov Models and deep learning that provide some evidence in this direction \citep{Milone2012-ap, Ferrero2023-vr}.

Regarding the difference between the values obtained in validation and test, except for options b) and c), some improvements are observed in the test performances. This may be caused because the amount of data with which the models are trained varies substantially, being 25\% larger in the test set. It is important to highlight that the test data set includes signals extracted from the same fieldwork, where the animals, equipment and experimental conditions were the same. If any of these conditions vary, the performance of the models may not be the same. This aspect is of special interest and \textcolor{black}{should be studied in the future}.

With respect to the inference times in Table~\ref{tab:tab6}, ten executions per method were run on the same hardware, and \textcolor{black}{the average and the SD of times} to predict 1 minute of signal are reported. 

In addition, the sum of FLOPs required to process one chunk from input signal (300 ms) is presented in this table. The same methodology reported in \cite{Ferrero2023-vr} was used to calculate these costs. From this comparison, it can be concluded that recurrent layers represent the biggest impact in terms of the processing time and operations of the proposed model, directly affected by the total number of parameters included in the block with RNNs layers. Inference times remained at similar levels without considerable differences between the proposed model and the rest of the options.

\section{Conclusions}
\label{S:6}

In this study, a multi-head CNN-RNN was introduced for JM event detection and recognition in grazing cattle. The model includes acoustic and IMU signals as inputs. The proposed architecture was compared with several different proposals among the three main data-level fusion strategies: data-level, feature-level, and decision-level. Variations in the number of layers, kernels, CNN heads, and kernel sizes were evaluated during the exploration. Additionally, different combinations of input signals were tested.

The results suggest that the proposed model for feature-level fusion is the more appropriate strategy in this context, using an independent CNN head for each input signal, achieving an average micro F1-score of 0.802. The contribution of each part of the model was also assessed and presented in an ablation study. Additionally, the effect of different window sizes was analysed, showing a clear advantage when using a size close to the average duration of the JM events (or even smaller). The proposed model clearly outperformed \textcolor{black}{the} state-of-the-art methods by at least 10\% (micro F1-score). 

This study pioneers research into the effectiveness of information fusion strategies for the detection and recognition of JM events in grazing cattle. The results demonstrate that the use of both sound and motion signals provides a clear advantage over unimodal solutions.

Despite the set of model architectures explored during experimentation, there are other potential changes that could be \textcolor{black}{beneficial}. The use of different fusion methods other than those used during experimentation, such as attention layers, will be evaluated in future works.

It is important to note several limitations of the presented model. From a practical point of view, its ability to generalize should be tested across different scenarios, including variations in recording devices, environmental conditions, \textcolor{black}{sensor placement}, different herd management, and even its applicability to other ruminants. Additionally, external noises can independently affect both signals, which may affect the model's performance. Future studies should investigate the robustness of the model and explore potential improvements in this aspect. \textcolor{black}{Ethical considerations related to sensor intrusiveness and animal welfare should be carefully evaluated before large-scale adoption. Future research should focus on overcoming these limitations to ensure the feasibility and scalability of AI-driven event detection in grazing cattle.}

Regarding technical limitations, the difficulty of obtaining \textcolor{black}{high-quality} labeled data sets with a larger amount of data \textcolor{black}{- which is a critical aspect in this context -} represents a challenge. The exploration of techniques to help overcome this problem, such as transfer learning or the use of semi-supervised approaches, will be evaluated in future research.

Finally, technical analysis and likely implications of the presented model are not covered in this study. However, one might notice that the information fusion imposes the need \textcolor{black}{for} both signals synchronization since mismatches can degrade the performance of the model. Another aspect related to the use of BGRU layers is the need of a buffer and, consequently, a potential problem for real-time applications. The model structure might \textcolor{black}{increase} resource demands in comparison with other approaches, which may limit implementation on edge devices. The use of knowledge distillation or model \textcolor{black}{pruning} while retaining performance might be an interesting line of research in this direction.

\section*{Acknowledgments}

This work has been funded by Universidad Nacional del Litoral, CAID 50620190100080LI and 50620190100151LI, Universidad Nacional de Rosario, projects 2013-AGR216, 2016-AGR266 and 80020180300053UR, Agencia San\-ta\-fe\-si\-na de Ciencia, Tecnología e Innovación (ASACTEI), project IO\--2018\-–00082, Consejo Nacional de Investigaciones Científicas y Técnicas (CONICET), project 2017-PUE sinc(i). Authors would like to thank the dedication and perceptive help by Campo Experimental J. Villarino Dairy Farm staff for their assistance and support during the completion of this study. Authors also gratefully acknowledge the support of NVIDIA Corporation with the donation of the Titan XP GPU used for this research. Finally, our thanks to Constanza Quaglia for her enormous contribution to this work through the development of the Android application used to capture signals during fieldwork.


\newpage
\bibliographystyle{apalike}
\bibliography{references.bib}







\end{document}